\documentclass{article}

\usepackage{amsmath,amsfonts,bm}

\def\eqref#1{equation~\ref{#1}}
\def\1{\bm{1}}

\DeclareMathAlphabet{\mathsfit}{\encodingdefault}{\sfdefault}{m}{sl}
\SetMathAlphabet{\mathsfit}{bold}{\encodingdefault}{\sfdefault}{bx}{n}

\DeclareMathOperator{\sign}{sign}

\usepackage{microtype}
\usepackage{graphicx}
\usepackage{subfigure}
\usepackage{booktabs}

\usepackage[resetlabels,labeled]{multibib}
\newcites{Supmat}{Supplementary References}

\newcommand{\fig}[1]{Figure~\ref{fig:#1}}
\newcommand{\sect}[1]{Section~\ref{sect:#1}}
\newcommand{\tab}[1]{Table~\ref{tab:#1}}

\newcommand{\eq}[1]{(\ref{eq:#1})}

\usepackage{hyperref}

\usepackage[accepted]{icml2020}

\icmltitlerunning{Unsupervised Discovery of Interpretable Directions in the GAN Latent Space}

\begin{document}

\twocolumn[
\icmltitle{Unsupervised Discovery of Interpretable Directions in the GAN Latent Space}

\begin{icmlauthorlist}
\icmlauthor{Andrey Voynov}{ya}
\icmlauthor{Artem Babenko}{ya,hse}
\end{icmlauthorlist}

\icmlaffiliation{ya}{Yandex, Russia}
\icmlaffiliation{hse}{National Research University Higher School of Economics
, Moscow, Russia}

\icmlcorrespondingauthor{Andrey Voynov}{an.voynov@yandex.ru}

\icmlkeywords{Generative Models}

\vskip 0.3in
]

\printAffiliationsAndNotice{}

\begin{abstract}
The latent spaces of GAN models often have semantically meaningful directions. Moving in these directions corresponds to human-interpretable image transformations, such as zooming or recoloring, enabling a more controllable generation process. However, the discovery of such directions is currently performed in a supervised manner, requiring human labels, pretrained models, or some form of self-supervision. These requirements severely restrict a range of directions existing approaches can discover. 

In this paper, we introduce an \textbf{unsupervised} method to identify interpretable directions in the latent space of a pretrained GAN model. By a simple model-agnostic procedure, we find directions corresponding to sensible semantic manipulations without any form of (self-)supervision. Furthermore, we reveal several non-trivial findings, which would be difficult to obtain by existing methods, e.g., a direction corresponding to background removal. As an immediate practical benefit of our work, we show how to exploit this finding to achieve competitive performance for weakly-supervised saliency detection. The implementation of our method is available online\footnote{\url{http://github.com/anvoynov/GanLatentDiscovery}}.
\end{abstract}

\section{Introduction}
\label{sect:intro}

Nowadays, generative adversarial networks (GANs) \cite{goodfellow2014generative} have become a leading paradigm of generative modeling in the computer vision domain. The state-of-the-art GANs \cite{big_gan, style_gan} are currently able to produce good-looking high-resolution images often indistinguishable from real ones. The exceptional generation quality paves the road to ubiquitous usage of GANs in applications, e.g., image editing \cite{isola2017image,zhu2017unpaired}, super-resolution \cite{ledig2017photo}, video generation \cite{wang2018video} and many others.

However, in most practical applications, GAN models are typically used as black-box instruments without a complete understanding of the underlying generation process. While several recent papers \cite{bau2019gandissect,voynov2019rpgan,yang2019semantic,style_gan,jahanian2019steerability,plumerault2019Controlling} address the interpretability of GANs, this research area is still in its preliminary stage.

\begin{figure}
    \centering
    \includegraphics[width=0.48\textwidth]{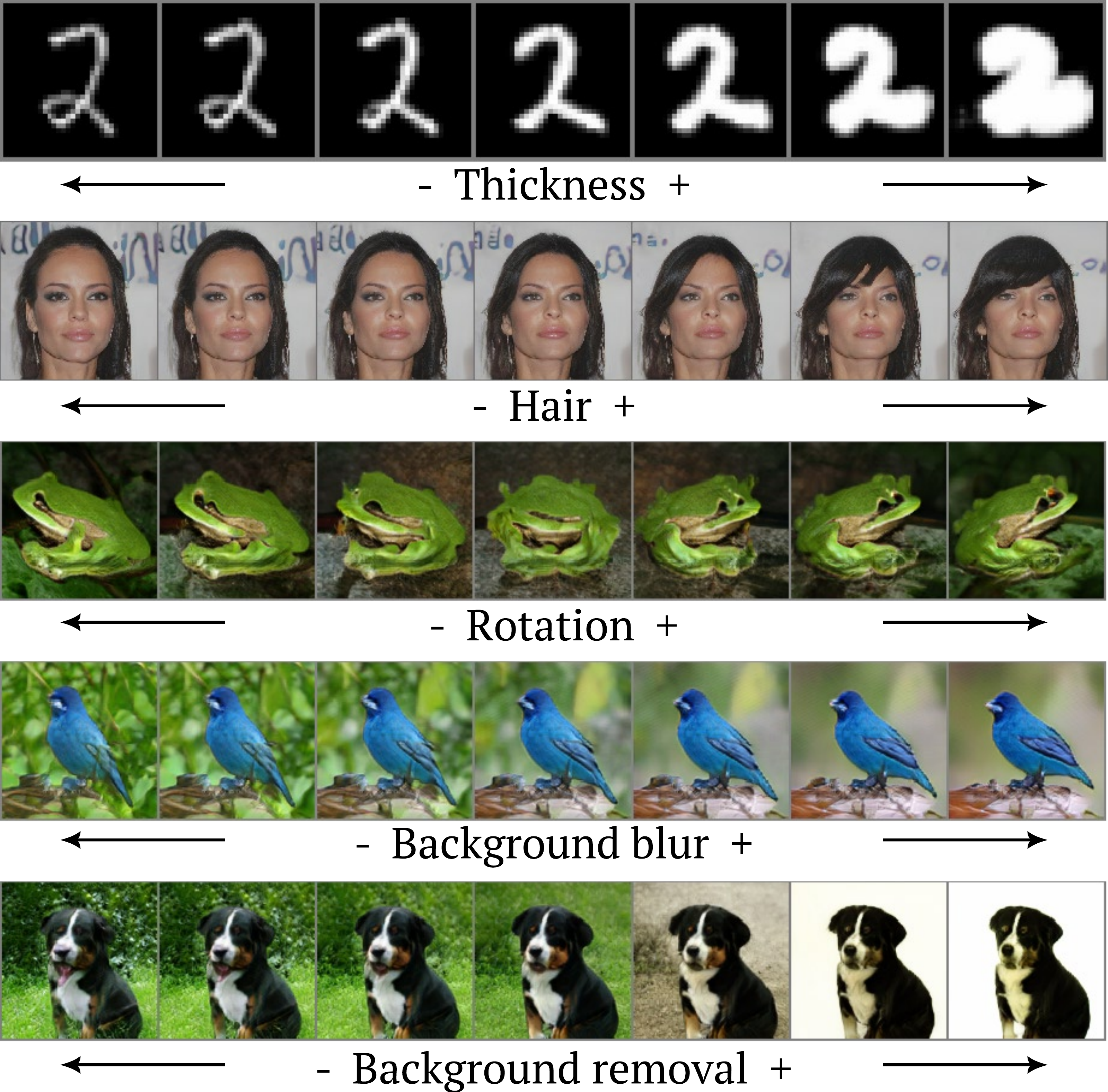}
    \vspace{-7mm}
    \caption{Examples of interpretable directions discovered by our unsupervised method for several datasets and generators.}
    \vspace{-7mm}
    \label{fig:tizer}
\end{figure}

An active line of study on GANs interpretability investigates the structure of their latent spaces. Namely, several works \cite{jahanian2019steerability,plumerault2019Controlling,goetschalckx2019ganalyze,shen2019interpreting} aim to identify semantically meaningful directions, i.e., corresponding to human-interpretable image transformations. At the moment, prior works have provided enough evidence that a wide range of such directions exists. Some of them induce domain-agnostic transformations, like zooming or translation \cite{jahanian2019steerability,plumerault2019Controlling}, while others correspond to domain-specific transformations, e.g., adding smile or glasses on face images \cite{radford2015unsupervised}.

While the discovery of interpretable directions has already been addressed by prior works, all these works require some form of supervision. For instance, \cite{shen2019interpreting, goetschalckx2019ganalyze, style_gan} require explicit human labeling or pretrained supervised models, which can be expensive or even impossible to obtain. Recent works \cite{jahanian2019steerability,plumerault2019Controlling} develop self-supervised approaches, but they are limited only to directions, corresponding to simple transformations achievable by automatic data augmentation.

In this paper, we propose a \textbf{completely unsupervised} approach to discover the interpretable directions in the latent space of a pretrained generator. In a nutshell, the approach seeks a set of directions corresponding to diverse image transformations, i.e., it is easy to distinguish one transformation from another. Intuitively, under such formulation, the learning process aims to find the directions corresponding to the independent factors of variation in the generated images. For several generators, we observe that many of the obtained directions are human-interpretable, see \fig{tizer}.

As another significant contribution, our approach discovers new practically important directions, which would be difficult to obtain with existing techniques. For instance, we discover the direction corresponding to the background removal, see \fig{tizer}. In the experimental section, we exploit it to generate high-quality synthetic data for saliency detection and achieve competitive performance for this problem in a weakly-supervised scenario. We expect that exploitation of other directions can also benefit other computer vision tasks in the unsupervised and weakly-supervised niches.

As our main contributions we highlight the following:
\begin{enumerate}
    \item We propose the first unsupervised approach for the discovery of semantically meaningful directions in the GAN latent space. The approach is model-agnostic and does not require costly generator re-training. 
    \item For several common generators, we managed to identify non-trivial and practically important directions. The existing methods from prior works are not able to identify them without expensive supervision.
    \item We provide an example of immediate practical benefit from our work. Namely, we show how to exploit the background removal direction for weakly-supervised saliency detection.
\end{enumerate}

The paper is organized as follows. \sect{related} discusses the relevant ideas from prior literature. \sect{method} formally describes our approach, \sect{experiments} reports the results and \sect{segmentation} applies our finding to the weakly-supervised saliency detection. In \sect{ablation} we ablate hyperparameters. \sect{conclusion} concludes the paper.
\section{Related work}
\label{sect:related}

In this section, we describe the relevant research areas and explain the scientific context of our study.

\textbf{Generative adversarial networks} \cite{goodfellow2014generative} currently dominate the generative modeling field. In essence, GANs consist of two networks -- a generator and a discriminator, which are trained jointly in an adversarial manner. The role of the generator is to map samples from the latent space distributed according to a standard Gaussian distribution to the image space. The discriminator aims to distinguish the generated images from the real ones. More complete understanding of the latent space structure is an important research problem as it would make the generation process more controllable.

\textbf{Interpretable directions in the latent space.} Since the appearance of earlier GAN models, it is known that the GAN latent space often possesses semantically meaningful vector space arithmetic, e.g., there are directions corresponding to adding smiles or glasses for face image data \cite{radford2015unsupervised}. Since exploitation of these directions would make image editing more straightforward, the discovery of such directions currently receives much research attention. A line of recent works \cite{goetschalckx2019ganalyze,shen2019interpreting,style_gan} employs explicit human-provided supervision to identify interpretable directions in the latent space. For instance, \cite{shen2019interpreting,style_gan} use the classifiers pretrained on the CelebA dataset \cite{liu2015faceattributes} to predict certain face attributes. These classifiers are then used to produce pseudo-labels for the generated images and their latent codes. Based on these pseudo-labels, the separating hyperplane is constructed in the latent space, and a normal to this hyperplane becomes a direction that captures the corresponding attribute. Another work \cite{plumerault2019Controlling} solves the optimization problem in the latent space that maximizes the score of the pretrained model, predicting image memorability. Thus, the result of the optimization is a direction corresponding to the increase of memorability. The crucial weakness of supervised approaches above is their need for human labels or pretrained models, which can be expensive to obtain. Two recent works \cite{jahanian2019steerability,plumerault2019Controlling} employ self-supervised approaches and seek the vectors in the latent space that correspond to simple image augmentations such as zooming or translation. While these approaches do not require supervision, they can be used to find only the directions capturing simple transformations that can be obtained automatically. 

All these approaches are able to discover only directions, which researchers expect to identify. In contrast, our unsupervised approach often identifies surprising directions, corresponding to non-trivial image manipulations.

\textbf{Disentanglement learning}. An alternative line of research on the model interpretability aims to train generators with disentangled latent spaces \cite{infogan, beta-VAE, OOGAN, lee2020high, ramesh2018spectral}. In particular, the seminal InfoGAN model \cite{infogan}
enforces the generated images to preserve information about the latent code coordinates by maximizing the corresponding mutual information. Another notable work proposes the $\beta$-VAE \cite{beta-VAE} model, which puts more emphasis on the $KL$-term in the standard VAE's ELBO objective. This objective modification requires the latent codes to be more ``efficient'', which is shown to result in disentangled representations. 

While these models do achieve disentanglement of their latent spaces, they are often inferior in terms of generation quality and diversity. Several recent papers address these issues by improving the original architectures and training protocols. For instance, \cite{OOGAN} forces the code vector $c$ to be one-hot, simplifying the task for a GAN discriminators' head to predict the code. The authors of \cite{lee2020high} combine VAE and GAN to achieve a disentanglement images representation by the VAE and then pass the discovered code to the GAN model. At the moment, it is unclear if disentangled generative models can be competitive to the state-of-the-art generators, e.g. BigGAN. In contrast, our method does not affect the pretrained generator distribution. %Another crucial difference is that we achieve interpretability and disentanglement by processing pairs of images. We force the images to have a latent space local shift and reconstruct it by comparison instead of full latent vector reconstruction. In a few words, we investigate interpretability by \textit{comparison} instead of \textit{reconstruction}.

\textbf{Jacobian decomposition}. Probably, the closest to ours is a recent work \cite{ramesh2018spectral} that also investigates the latent space of a pretrained GAN model. They note that the left eigenvectors of the generators' Jacobian matrix can serve as the most disentangled directions. The authors also propose an iterative algorithm that constructs an ``interpretable curve'' starting from a latent point $z_0$ and moving it in a direction of the Jacobians' $k$-th left eigenvector at each point. Once the latent vector moves along that curve, the generated image appears to be transformed by a human-meaningful transformation. Nevertheless, while the constructed curves often capture interpretable transformations, their effects are typically entangled (i.e. lighting and geometrical transformations appear simultaneously). This method also requires an expensive (in terms of both memory and runtime) iterative process computing the Jacobian matrix on each step of the curve construction and has to be applied for each latent code independently. On the contrary, we propose a lightweight approach that identifies a set of the directions at once. The method from \cite{ramesh2018spectral} is also limited with the maximal number of discovered directions equal to the latent space dimensionality,  while our approach can be applied for a higher number of directions.
\section{Method}
\label{sect:method}

\subsection{Motivation}

\begin{figure}[h!]
    \centering
    \includegraphics[width=0.45\textwidth]{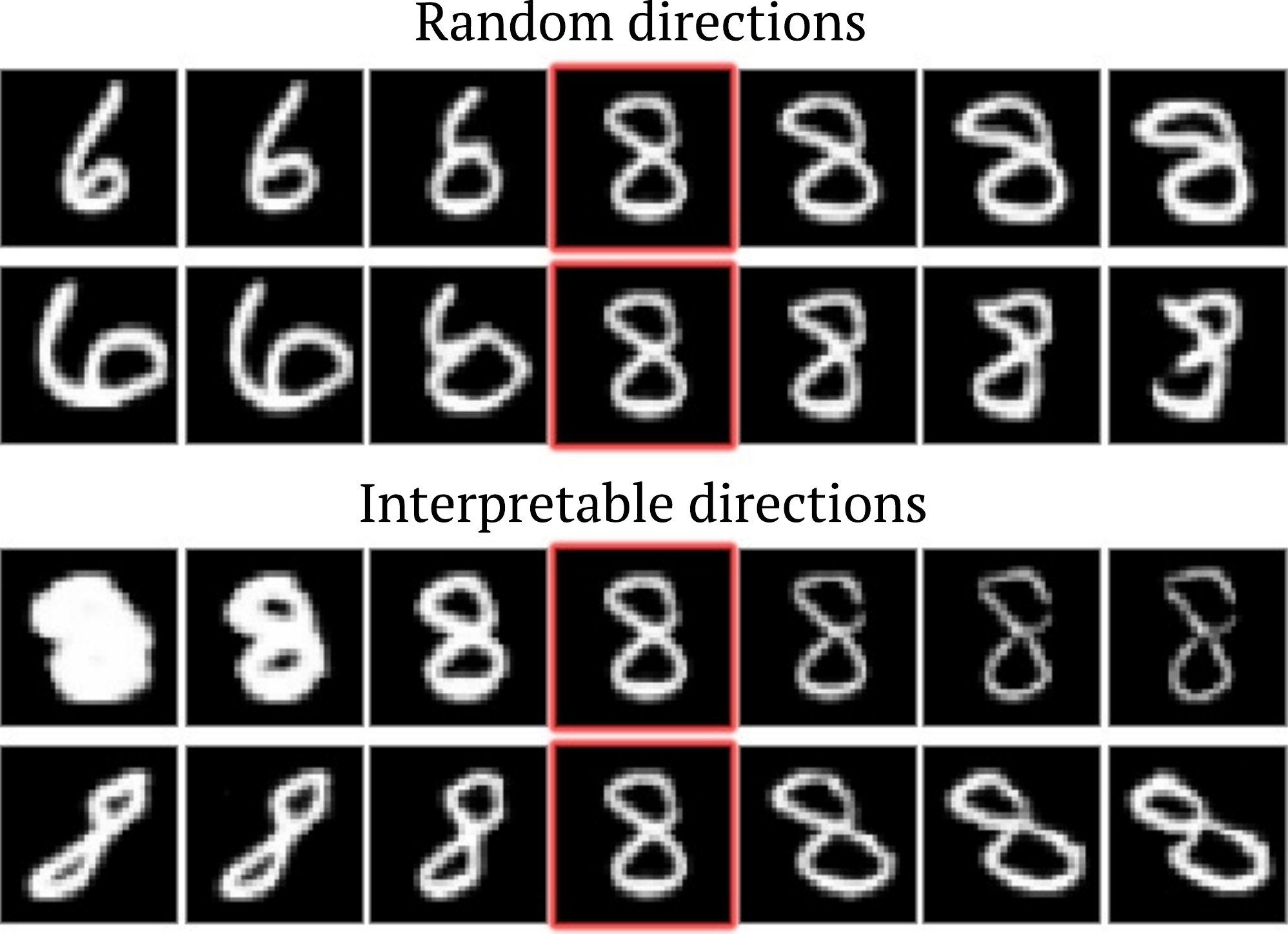}
    \vspace{-3mm}
    \caption{Image transformations obtained by moving in random (top) and interpretable (bottom) directions in the latent space.}
    \label{fig:motivation}
\end{figure}

Before a formal description of our method, we explain its underlying motivation by a simple example on \fig{motivation}.
\fig{motivation} (top) shows the transformations of an original image (in a red frame) obtained by moving in two random directions of the latent space for the Spectral Norm GAN model \cite{miyato2018spectral} trained on the MNIST dataset \cite{MNIST}. As one can see, moving in a random direction typically affects several factors of variations at once, and different directions ``interfere'' with each other. This makes it difficult to interpret these directions or to use them for semantic manipulations in image editing.

The observation above provides the main intuition behind our method. Namely, we aim to learn a set of directions inducing ``orthogonal'' image transformations that are easy to distinguish from each other. We achieve this via jointly learning a set of directions and a model to distinguish the corresponding image transformations. The high quality of this model implies that directions do not interfere; hence, hopefully, affect only a single factor of variation and are easy-to-interpret.

\subsection{Learning}

\begin{figure*}
    \centering
    \includegraphics[width=0.999\textwidth]{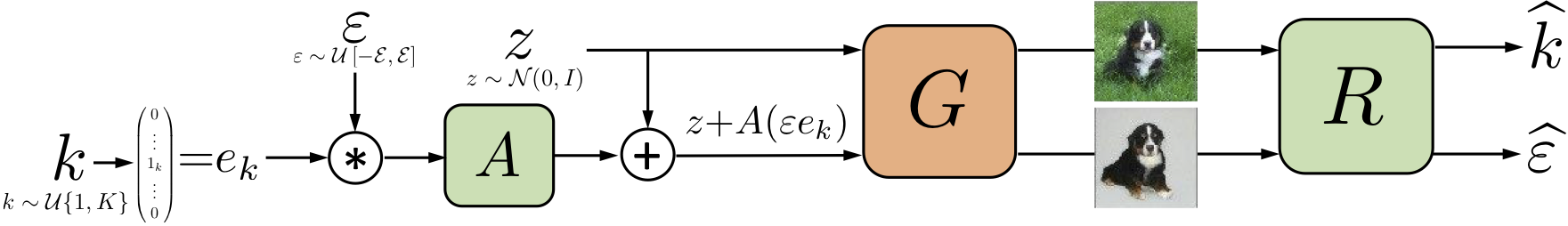}
    \caption{Scheme of our learning protocol, which discovers interpretable directions in the latent space of a pretrained generator $G$. A training sample in our protocol consists of two latent codes, where one is a shifted version of another. Possible shift directions form a matrix $A$. Two codes are passed through $G$ and an obtained pair of images go to a reconstructor $R$ that aims to reconstruct a direction index $k$ and a signed shift magnitude $\varepsilon$.}
    \label{fig:rectification}
    \vspace{-2mm}
\end{figure*}

The learning protocol is schematically presented on \fig{rectification}. Our goal is to discover the interpretable directions in the latent space of a pretrained GAN generator $G:z\longrightarrow I$, which maps samples from the latent space $z \in \mathbb{R}^d$ to the image space. $G$ is a non-trainable component of our method, and its parameters do not change during learning.
Two trainable components of our method are:
\begin{enumerate}
    \item A matrix $A \in \mathbb{R}^{d \times K}$, where $d$ equals to the dimensionality of the latent space of $G$. A number of columns $K$ determines the number of directions our method will discover. It is a hyperparameter of our method, and we discuss its choice in the next section. In essence, the columns of $A$ correspond to the directions we aim to identify.
    \item A \textit{reconstructor} $R$, which obtains an image pair of the form $(G(z),\ G(z+A(\varepsilon  e_k)))$, where the first image is generated from a latent code $z \sim \mathcal{N}(0,I)$, while the second one is generated from a \textit{shifted} code $z{+}A(\varepsilon  e_k)$. Here $e_k$ denotes an axis-aligned unit vector $(0,\dots,1_k,\dots,0)$ and $\varepsilon$ is a scalar. In other words, the second image is a transformation of the first one, corresponding to moving by $\varepsilon$ in a direction, defined by the $k$-th column of $A$ in the latent space.
    The reconstructor's goal is to reproduce the shift in the latent space that induces a given image transformation. In more details, $R$ produces two outputs $R(I_1,I_2) = \left(\widehat{k}, \widehat{\varepsilon}\right)$, where $\widehat{k}$ is a prediction of a direction index $k \in \{1,\dots,K\}$, and $\widehat{\varepsilon}$ is a prediction of a shift magnitude $\varepsilon$. More formally, the reconstructor performs a mapping $R:(I_1,\ I_2)\longrightarrow(\{1,\dots,K\},\ \mathbb{R})$. 
\end{enumerate}

\textbf{Optimization objective.} Learning is performed via minimizing the following loss function:

\begin{equation}
\begin{split}
\underset{A,R}{\text{min}} \hspace{2mm} \underset{z,k,\varepsilon}{\mathbb{E}}L(A,R) = \underset{A,R}{\text{min}}\hspace{2mm}\underset{z,k,\varepsilon}{\mathbb{E}}\left[L_{cl}(k,\widehat{k}) + \lambda L_r (\varepsilon, \widehat{\varepsilon})\right]
\end{split}
\label{eq:L_total}
\vspace{-2mm}
\end{equation}

For the classification term $L_{cl}(\cdot,\cdot)$ we use the cross-entropy function, and for the regression term $L_{r}(\cdot,\cdot)$ we use the mean absolute error. In all our experiments we use a weight coefficient $\lambda{=}0.25$.

As $A$ and $R$ are optimized jointly, the minimization process seeks to obtain such columns of $A$ that the corresponding image transformations are easier to distinguish from each other, to make the classification problem for reconstructor simpler. In the experimental section below, we demonstrate that these ``disentangled'' directions often appear to be human-interpretable.

\begin{figure}[h!]
    \centering
    \includegraphics[width=0.50\textwidth]{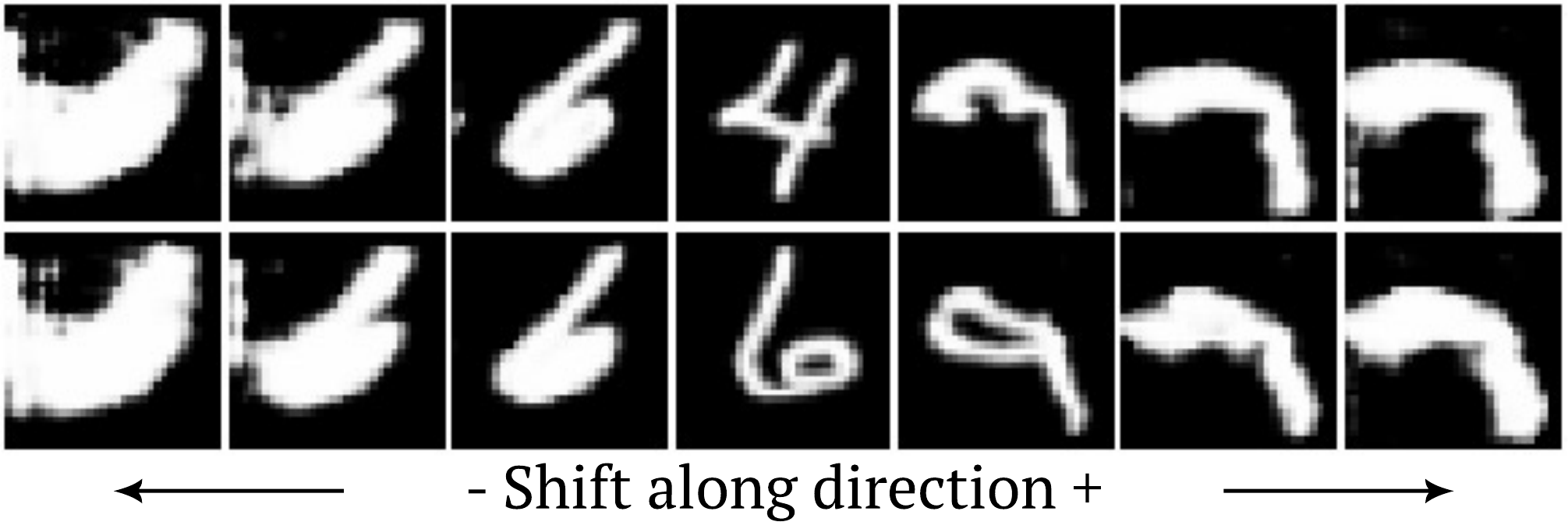}
    \vspace{-3mm}
    \caption{Direction of collapsing variation. The regression term in our objective prevents discovery of such directions.}
    \vspace{-3mm}
    \label{fig:collapse}
\end{figure}

The role of the regression term $L_r$ is to force shifts along discovered directions to have the continuous effect, thereby preventing ``abrupt'' transformations, e.g., mapping all the images to some fixed image. See \fig{collapse} for a latent direction example that maps all the images to a fixed one.

\subsection{Practical details.}

Here we describe the practical details of the pipeline and explain our design choices.

\textbf{Reconstructor architecture.} For reconstructor models $R$ we use the LeNet backbone \cite{lenet} for the MNIST and AnimeFaces and the ResNet-$18$ model \cite{resnet} for Imagenet and CelebA-HQ. In all experiments, a number of input channels is set to six (two for MNIST), since we concatenate the input image pair along channels dimension. We also use two separate ``heads'', predicting a direction index and a shift magnitude, respectively.

\textbf{Distribution of training samples.} The latent codes $z$ are always sampled from the normal distribution $\mathcal{N}(0, I)$ as in the original $G$. The direction index $k$ is sampled from a uniform distribution $\mathcal{U}\{1, K\}$. A shift magnitude $\varepsilon$ is sampled from the uniform distribution $\mathcal{U}[-6, 6]$. We also found a minor advantage in forcing $\varepsilon$ to be separated from $0$ as too small shifts almost do not affect the generated image. Thus, in practice after sampling we take $\varepsilon$ equal to $\sign(\varepsilon) \cdot \max(|\varepsilon|, 0.5)$. We did not observe any difference from using other distributions for $\varepsilon$.

\textbf{Choice of $K$.} The number of directions $K$ is set to be equal to the latent space dimensionality for Spectral Norm GAN \cite{miyato2018spectral} with Anime Faces dataset and BigGAN \cite{big_gan} (which are $128$ and $120$). For Spectral Norm GAN with MNIST dataset, we use  $K{=}64$, since its latent space is $128$-dimensional, and it is too difficult for our model to obtain so many different interpretable directions for simple digit images. Following the same considerations for ProgGAN \cite{karras2017progressive} we use $K{=}200$ since its latent space is $512$-dimensional.

\textbf{Choice of $A$.} We experimented with three options for $A$:%: \mathbb{R}^K \longrightarrow \mathbb{R}^d$:

\begin{itemize}
    \item $A$ is a general linear operator;
    \item $A$ is a linear operator with all matrix columns having a unit length;
    \item $A$ is a linear operator with orthonormal matrix columns.
\end{itemize}

The first option appeared to be impractical as during the optimization, we frequently observed the columns of $A$  to have very high $l_2$-norms. The reason is that for a constant latent shift $c$ of a high norm, most of the generated samples $G(z + c)$ with $z \sim \mathcal{N}(0, I)$ appears to be almost the same for all $z$. Thus the classification term in the loss \eq{L_total} pushes $A$ to have columns of a high norm to simplify classification. 

In all experiments, we use either $A$ with columns of length one (the second option) either with orthonormal columns (the third option). To guarantee that $A$ has unit-norm columns, we divide each column by its length. For the orthonormal case, we parametrize $A$ with a skew-symmetric matrix $S$ (that is $S^T{=}-S$) and define $A$ as the first $K$ columns of the exponent of $S$, see details in supplementary. %See details in \sect{ortho_sup}. % In that case, we also suppose that $K \leq d$.

%\footnote{In fact the map $\exp : \mathrm{Skew}_d \to SO(d)$ yields the operators with determinant 1. This is not a limitation as we take signed shift multiplicator $\varepsilon$}

In experiments, we observe that both two options discover similar sets of interpretable directions. In general, using matrix $A$ with unit-norm columns is more expressive and is able to find more directions. However, on some datasets, the option with orthonormal columns discovered more interesting directions. In the experiments, we use orthonormal $A$ for AnimeFaces and BigGAN and unit length columns for MNIST and ProgGAN.

\section{Experiments}
\label{sect:experiments}

Here we evaluate our approach on several datasets in terms of both quantitative and qualitative results. In all experiments, we do not exploit any form of external supervision and operate in a completely unsupervised manner.

\textbf{Datasets and generator models.} We experiment with four common datasets and generator architectures:

\begin{enumerate}
    \item MNIST \cite{MNIST}, containing $32\times32$ images. Here we use Spectral Norm GAN \cite{miyato2018spectral} with ResNet-like generator of three residual blocks.
    
    \item AnimeFaces dataset \cite{anime_faces}, containing $64\times64$ images. For AnimeFaces we use Spectral Norm GAN \cite{miyato2018spectral} with ResNet-like generator of four residual blocks.

    \item CelebA-HQ dataset \cite{liu2015faceattributes}, containing $1024 \times 1024$ images. We use a pretrained ProgGAN generator \cite{karras2017progressive}, available online\footnote{\url{http://github.com/ptrblck/prog_gans_pytorch_inference}}.
    
    \item BigGAN generator \cite{big_gan} trained on ILSVRC dataset \cite{imagenet_cvpr09}, containing $128 \times 128$ images. We use the BigGAN, available online\footnote{\url{http://github.com/ajbrock/BigGAN-PyTorch}}.
\end{enumerate}

\textbf{Optimization.} In all the experiments, we use the Adam optimizer to learn both the matrix $A$ and the reconstructor $R$. We always train the models with a constant learning rate $0.0001$. We perform $2 \cdot 10^5$ gradient steps for ProgGAN and $10^5$ steps for others as the first has a significantly higher latent space dimension. We use a batch size of $128$ for Spectral Norm GAN on the MNIST, and Anime Faces datasets, a batch size of $32$ for BigGAN, and a batch size of $10$ for ProgGAN. All the experiments were performed on the NVIDIA Tesla v100 card.

\textbf{Evaluation metrics.} Since it is challenging to measure interpretability and disentanglement directly, we propose two evaluation measures described below.

\begin{enumerate}
    \item \textit{Reconstructor Classification Accuracy (RCA)}. As described in \sect{method}, the reconstructor $R$ aims to predict what direction in the latent space produces a given image transformation. In essence, the reconstructor's classification ``head'' solves a multi-class classification problem. Therefore, high RCA values imply that directions are easy to distinguish from each other, i.e., corresponding image transformations do not ``interfere'' and influence different factors of variations. While it does not mean interpretability directly, in practice, transformations affecting a few factors of variation are easier to interpret. RCA allows us to compare the directions obtained with our method with random directions or with directions corresponding to coordinate axes. To obtain RCA values for random or standard coordinate directions, we set $A$ to be equal random or identity matrix and do not optimize it during learning.

    \item \textit{Individual interpretability (mean-opinion-score, MOS)}. To quantify the interpretability of individual directions, we perform human evaluation. For assessment, we employ eleven human assessors, all having ML/CV background. The evaluation protocol is the following:
    \begin{itemize}
        \item[-] For each assessor we sample ten random $z \sim \mathcal{N}(0,I)$;
        \item[-] For each direction $h$ we plot a chart similar to \fig{mnist_directions}. Namely, we plot $G(z + s \cdot h)$, varying $s$ from $-8$ to $8$ for all $z$ sampled on the previous step).
    \end{itemize}
    The assessor is then asked two questions:
    \begin{itemize}
        \item[-] Does $h$ operate consistently for different $z$?
        \item[-] Does $h$ affect a single factor of variation, which is easy-to-interpret?
    \end{itemize}
    If $h$ meets both requirements, it is treated as ``interpretable'' and is marked as $1$. Otherwise it is marked as $0$. To obtain a final MOS value for a set of directions, we average the marks across all assessors and all directions from the set. For a fair comparison, we evaluate different sets of directions on completely the same $z$.
\end{enumerate}

While MOS measures the quality of directions independently, high RCA values indicate that discovered directions are substantially different, so both metrics are important. Therefore, we report MOS and RCA for directions discovered by our method for all datasets. We compare to directions corresponding to coordinate axes and random orthonormal directions in \tab{quantitative}. Along with quantitative comparison, we provide the qualitative results for each dataset below.

\subsection{MNIST}

\begin{figure}
    \centering
    \includegraphics[width=0.49\textwidth]{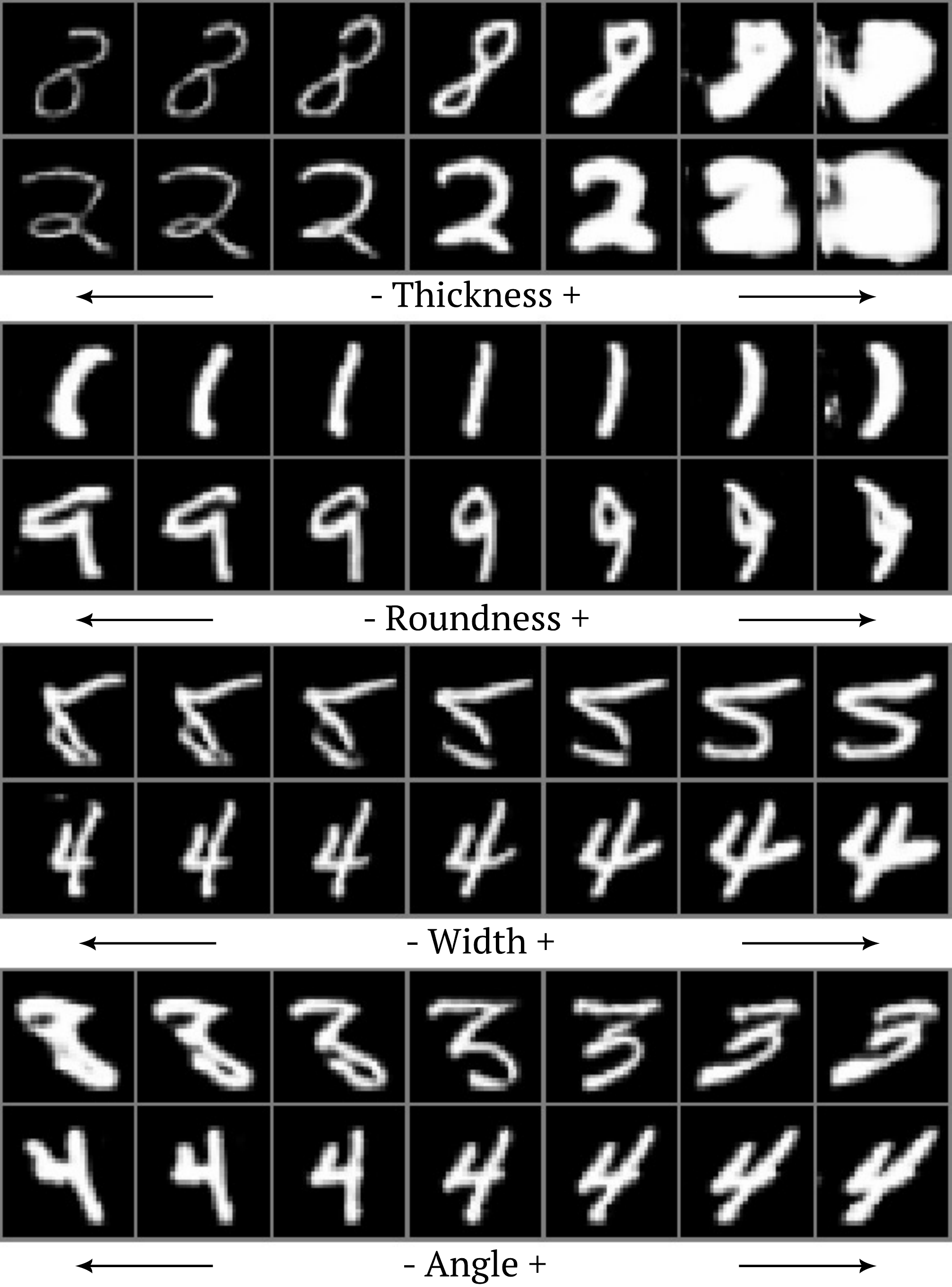}
    \vspace{-3mm}
    \caption{Examples of interpretable directions for Spectral Norm GAN and MNIST.}
    \vspace{-3mm}
    \label{fig:mnist_directions}
\end{figure}

Qualitative examples of transformations induced by directions obtained with our method are shown on \fig{mnist_directions}. The variations along learned directions are easy to interpret and transform all samples in the same manner for any $z$.

% Quantitatively, RCA values for random directions and directions corresponding to coordinate axes equal $0.33$ and $0.34$, respectively. Meanwhile, RCA for learned directions is $0.77$, which confirms the advantage of learnable directions. DVN --- TODO.
%As the Spectral Norm GAN with a ResNet-like backbone has no hierarchy in the latent space, all variations along its latent space coordinates are typically looks like the top two rows on \fig{motivation}. 

% First, to quantify the effectiveness of the proposed method, we compare the accuracy of the latent direction classifier $R$ with $A = \mathrm{id}$ and the proposed learnable $A$ with unit columns, optimized simultaneously with $R$. That is, we compare the tasks of initial axis shifts classification with the discovered directions shifts classification. In the first case the classifier $R$ performs with the accuracy $\approx 0.33$ while in the second case its accuracy is equal to $\approx 0.77$. Thus, we conclude that the found directions induce more various and persistent generated image transforms. Notably, in all other experiments the accuracy of the classifier takes values above $0.9$. Apparentely, this happens due to a low level of images variability in MNIST compared to other domains.

\begin{figure}
    \centering
    \includegraphics[width=0.49\textwidth]{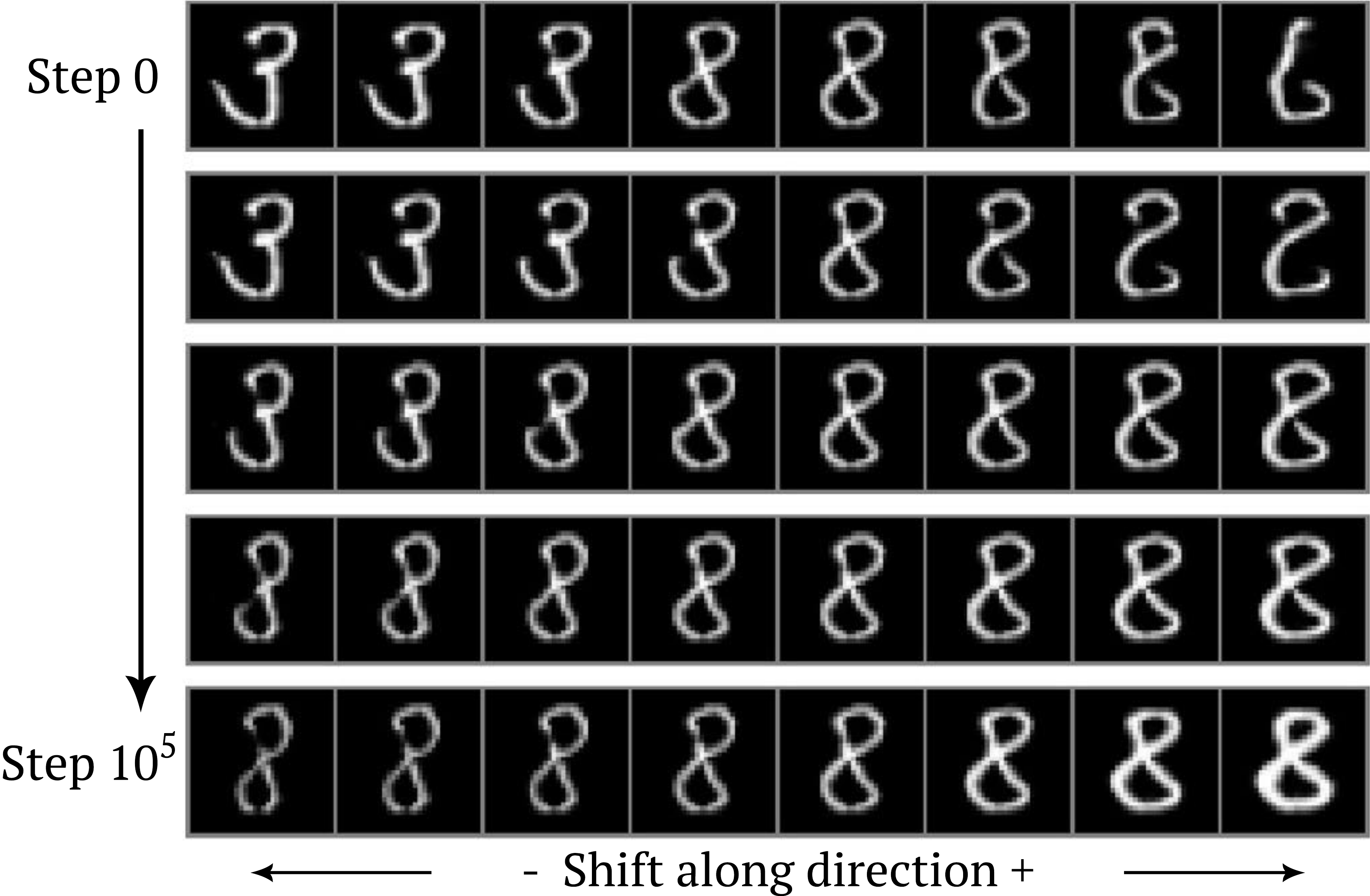}
    \vspace{-3mm}
    \caption{Image variation along a particular column of $A$ during the optimization process. Before optimization, the corresponding transformation affects several factors of variation, and gradually “concentrates” only on the digit thickness as optimization proceeds.
}
    \vspace{-3mm}
    \label{fig:mnist_evolution}
\end{figure}

\textbf{Evolution of directions.} On \fig{mnist_evolution} we illustrate how the image variation along a given direction evolves during the optimization process. Namely, we take five snapshots of the matrix $A$: $A_{step = 0}, \dots A_{step = 10^5}$ from different optimization steps. Hence $A_{step = 0}$ is the identity transformation and $A_{step = 10^5}$ is the final matrix of directions. Here we fix a direction index $k$ and latent $z \in \mathbb{R}^{128}$. The $i$-th row on \fig{mnist_evolution} are the images $G(z + A_{step = 25 \cdot 10^3 \cdot (i - 1)}(\varepsilon \cdot e_k))$. As one can see, in the course of optimization the direction stops to affect digit type and ``focuses'' on thickness.

 \begin{figure}[!t]
    \centering
    \includegraphics[width=0.48\textwidth]{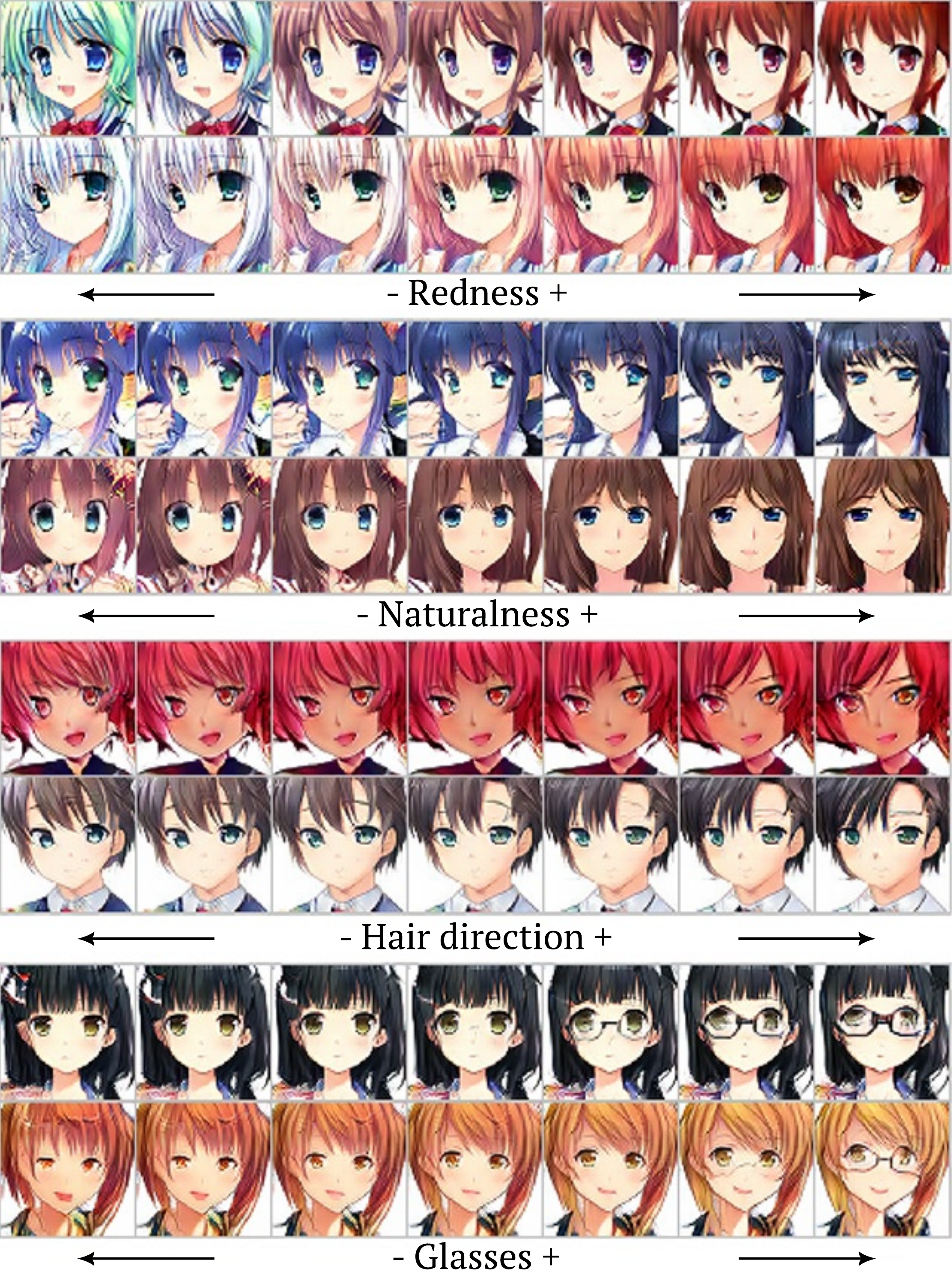}
    \vspace{-3mm}
    \caption{Examples of directions discovered for Spectral Norm GAN and AnimeFaces dataset.}
    \vspace{-3mm}
    \label{fig:anime_directions}
\end{figure}

\subsection{Anime Faces}
On this dataset, we observed advantage of orthonormal $A$ compared to $A$ with unit-norm columns. We conjecture that the requirement of orthonormality can serve as a regularization, enforcing diversity of directions. However, we do not advocate the usage of orthonormal $A$ for all data since it did not result in practical benefits for MNIST/CelebA.
On the \fig{anime_directions}, we provide examples discovered by our approach.
%The RCA values for random, coordinate, and learned directions are TODO, TODO, and TODO, respectively, which confirms the success of our method. DVN - TODO.
%On the \fig{anime_evolution}, similarly to \fig{mnist_evolution}, we depict an optimized direction shifts evolution during the training.
 
\subsection{ProgGAN}
\begin{figure}
    \centering
    \includegraphics[width=0.48\textwidth]{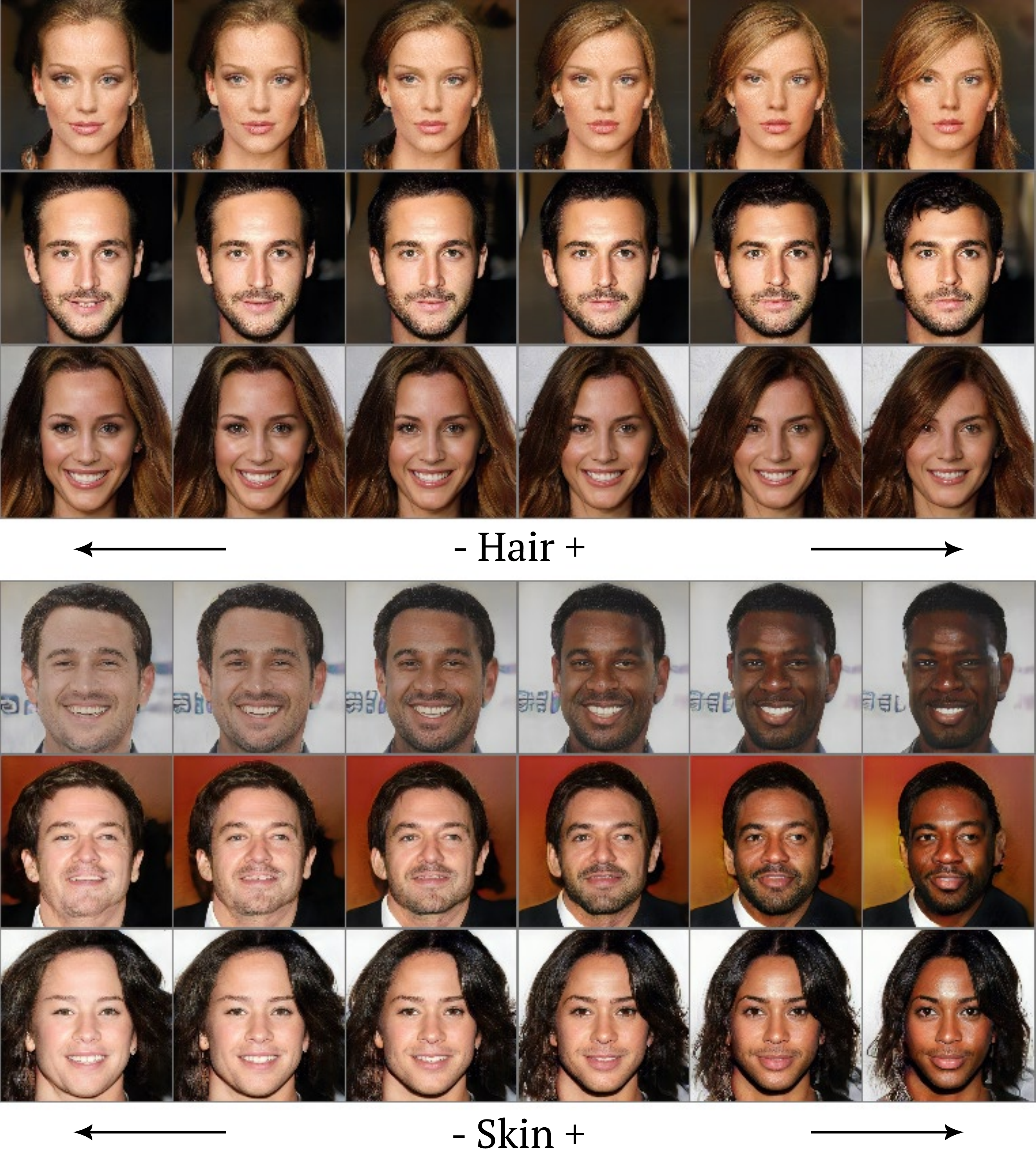}
    \vspace{-3mm}
    \caption{Examples of directions discovered for ProgGAN and CelebA dataset.}
    \label{fig:proggan_directions}
\end{figure}

Since the latent space dimensionality for ProgGAN equals $512$ and is remarkably higher compared to other models, we observed that the reconstructor RCA values notably degrade with $K{=}512$ and we set $K{=}200$ in this experiment.
See \fig{proggan_directions} for examples of discovered directions for ProgGAN. These directions are likely to be useful for face image editing and are challenging to obtain without supervision.% The RCA values for $K{=}200$ random, coordinate, and learned directions are TODO, TODO, and TODO respectively. DVN - TODO.

\subsection{BigGAN}

\begin{figure}[!t]
    \centering
    \includegraphics[width=0.48\textwidth]{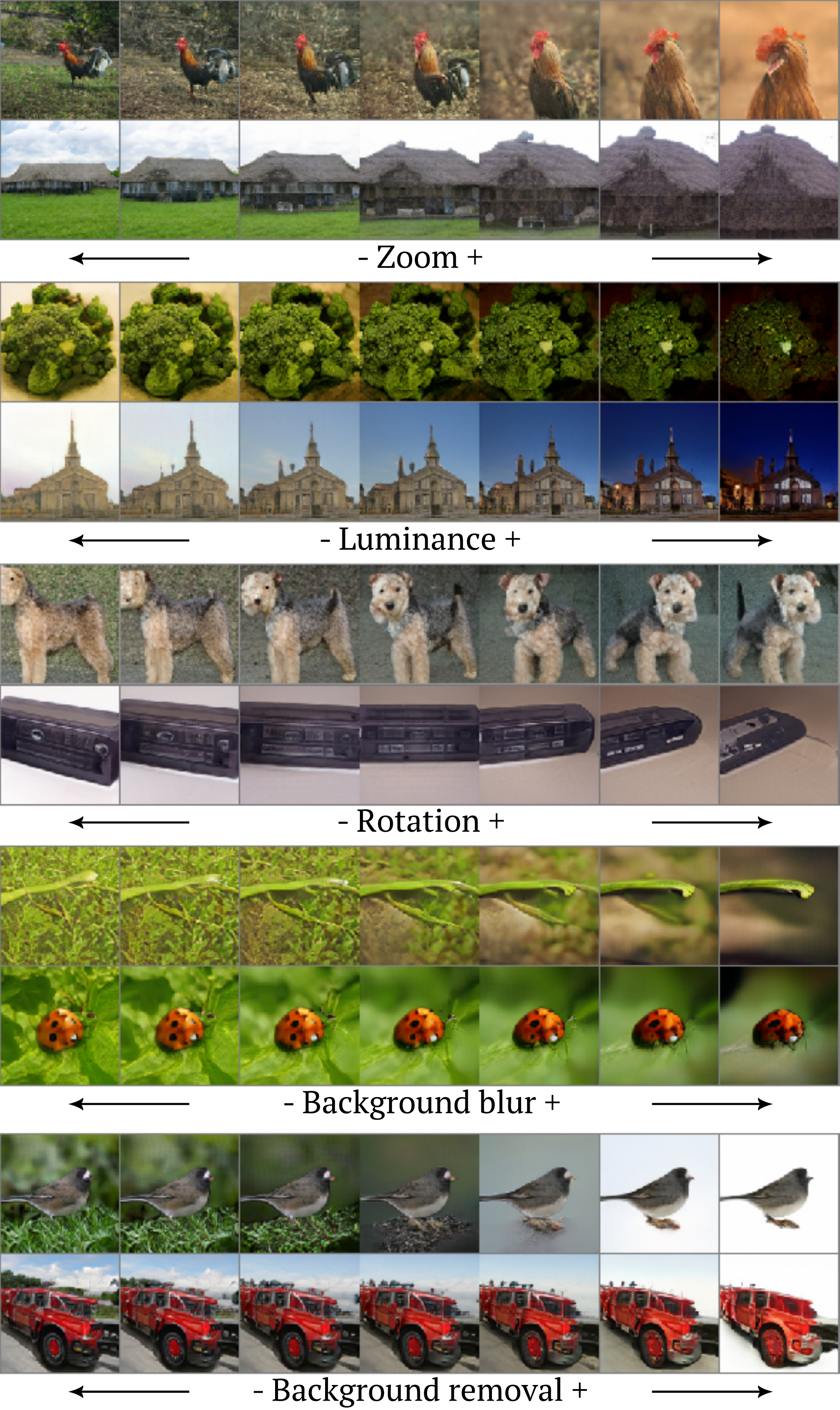}
    \vspace{-3mm}
    \caption{Examples of directions discovered for BigGAN.}
    \vspace{-0mm}
    \label{fig:biggan_directions}
\end{figure}

Several examples of directions discovered by our method are presented on \fig{biggan_directions}. In this dataset, our method reveals several interesting directions, which can be of significant practical importance. For instance, we discover directions, corresponding to background blur and background removal, which can serve as a valuable source of training data for various computer vision tasks, as we show in the following section. Here we also use orthonormal $A$ since it results in a more diversified set of directions.

For BigGAN we also perform more detailed analysis by asking the assessors to categorize the interpretable directions into three types:
\begin{itemize}
    \item Geometry (e.g. zoom / shift / rotation);
    \item Texture (e.g. background blur / add grass / sharpness);
    \item Color (e.g. lighting / saturation).
\end{itemize}
Results are presented in \tab{biggan_cmp_details}.
Notably, interpretable coordinate directions mostly belong to the color or texture types, while interpretable random directions mostly affect geometry (all corresponding to zooming).

\begin{table}[]
    \centering
    \begin{tabular}{|c||c|c|c|}
        \hline
         Category $\backslash$ Direction & Random & Coordinate & Ours \\
         \hline
         Geometry & 0.84 & 0.17 & 0.45\\
         \hline
         Coloring & 0.16 & 0.45 & 0.2\\
         \hline
         Textural & 0 & 0.38 & 0.35\\
         \hline
    \end{tabular}
    \caption{Rates of different types of transformations among interpretable directions in the BigGAN latent space.}
    \label{tab:biggan_cmp_details}
\end{table}

%In \sect{dvn_sup} we discuss an alternative for Individual interpretability evaluation that does not involve a human supervision.

% for some of the discovered BigGAN interpretable directions and \fig{biggan_evolution} for an optimized direction shifts evolution during the training.

% As the BigGAN generator model has skip-connections from input noise to intermediate layers, there is a chance that initial axis hold an informative variation direction. This is actually the case as most of them do affect image coloring. To handle the potential bias induced by these axis, we initialize $A$ with columns formed by uniformly distributed unit vectors in the unit-columns case and with the uniform distribution on $SO(d)$ in orthonormal case. Notably the directions discovered by our method lie far from the unit axis, yet some of them deal with a coloring modification. For instance, for the direction $h_{\mathrm{bg}}$ responsible for a background removal the closest axis direction satisfies $\max_i|\left<h_{\mathrm{bg}}, e_i\right>| \approx 0.25$ with angle $\approx 75^{\circ}$.
% Surprisingly we have found that the directions discovered by the method for a constant-class conditioned BigGAN also works well for other classes. We leave this intrigues notion for future works.

% See \fig{biggan_directions} for some of discovered BigGAN interpretable directions and \fig{biggan_evolution} for an optimized direction shifts evolution during the training.

\section{Weakly-supervised saliency detection}
\label{sect:segmentation}

In this section, we provide a simple example of practical usage of directions discovered by our method. Namely, we describe a straightforward way to exploit the background removal direction $h_{\mathrm{bg}}$ from the BigGAN latent space for a problem of weakly supervised saliency detection. In a nutshell, this direction can be used to generate high-quality synthetic data for this task. Below we always explicitly specify an Imagenet class passed to the BigGAN generator, i.e., $G(z, c),\  1 \leq c \leq 1000$.

\begin{figure}
    \centering
    \includegraphics[width=0.49\textwidth]{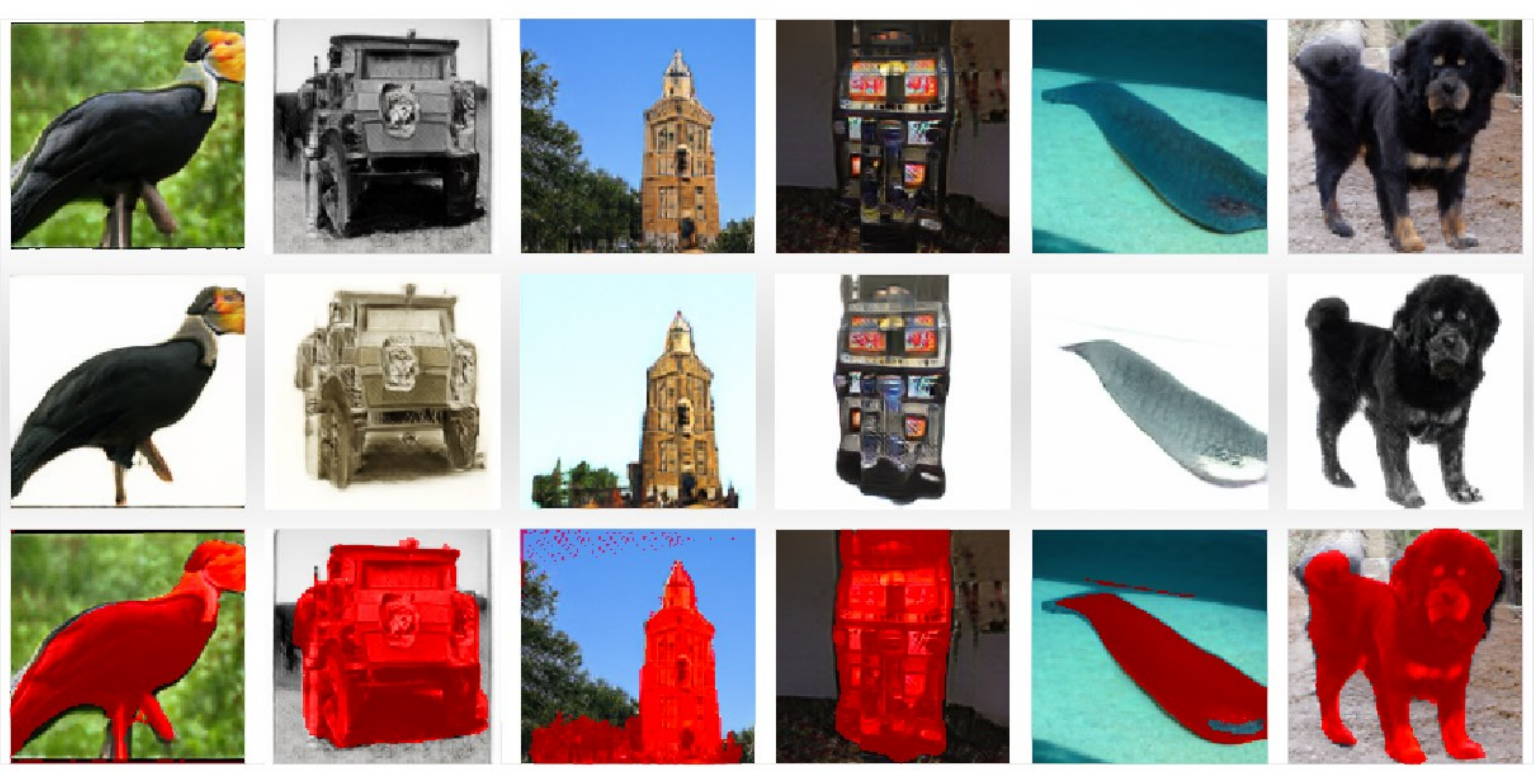}
    \vspace{-5mm}
    \caption{Segmentation masks for BigGAN samples used to train a saliency detection model. \textit{Line 1:} original samples $G(z)$; \textit{Line 2:} samples with reduced background $G(z + h_{\mathrm{bg}})$; \textit{Line 3:} generated binary masks obtained by thresholding.}
    \vspace{-3mm}
    \label{fig:segmentation_gen}
\end{figure}

\fig{biggan_directions} shows that $h_{\mathrm{bg}}$ is responsible for the background opacity variation. After moving in this direction, the pixels of a foreground object remain unchanged, while the background pixels become white. Thus, for a given BigGAN sample $G(z, c)$, one can label the white pixels from the corresponding shifted image $G(z + h_{\mathrm{bg}}, c)$ as background, see \fig{segmentation_gen}. Namely, to produce labeling, we compare an average intensity over three color channels for the image $G(z + h_{\mathrm{bg}}, c)$ to a threshold $\theta$:

\begin{equation}
\mathrm{Mask}(G(z, c)) = \left[\ G(z + h_{\mathrm{bg}}, c) < \theta\ \right]
\end{equation}

Assuming that intensity values lie in the range $[0, 1]$, we set $\theta = 0.95$.

Given such synthetic masks, it is possible to train a model that achieves high quality on real data. Let us have an image dataset $\mathcal{D}$. Then one can train a binary segmentation model on the samples $\left[G(z, c),\ \mathrm{Mask}(G(z, c))\right]$ with classes $c$ that frequently appear in images from $\mathcal{D}$. While the images of $\mathcal{D}$ can be unlabeled, we perform the following trick. We take an off-the-shelf pretrained Imagenet classifier (namely, ResNet-18) $M$. For each sample $x \in \mathcal{D}$ we consider five most probable classes from the prediction $M(x)$. Thus, for each of 1000 ILSVRC classes, we count a number of times it appears in the top-5 prediction over $\mathcal{D}$. Then we define a subset of classes $\mathcal{C}_\mathcal{D}$ as the top $25\%$ most frequent classes. 
Finally, we form a pseudo-labeled segmentation dataset with the samples $\left[G(z, c),\ \mathrm{Mask}(G(z, c))\right]$ with $z \sim \mathcal{N}(0, I), c \in \mathcal{C}_\mathcal{D}$. We exclude samples with mask area below $0.05$ and above $0.5$ of the whole image area. Then we train a segmentation model $U$ on these samples and apply it to the real data $\mathcal{D}$.

Note that the only supervision needed for the saliency detection method described above is image-level ILSVRC class labels. Our method does not require any pixel-level or dataset-specific supervision.

% \begin{table}
% \centering
% \caption{Quantitave comparison of our method with random and coordinate axes directions in terms of RCA and DVN.}
% \vspace{1mm}
%     \begin{tabular}{ |c|c|c|c|c| }
%         \hline
%         Directions & MNIST & Anime & CelebA & ILSVRC\\
%         \hline
%         \multicolumn{5}{|c|}{\bf RCA}\\
%         \hline
%         Random & 0.46 & 0.85 & 0.6 & 0.76\\ 
%         Coordinate & 0.48 & 0.89 & 0.82 & 0.66\\
%         Ours & \textbf{0.88} & \textbf{0.99} & \textbf{0.9} & \textbf{0.85}\\
%         \hline
%         \multicolumn{5}{|c|}{\bf DVN}\\
%         \hline
%         Random & 0.87 & 0.81 & 0.56 & 0.71\\ 
%         Coordinate & 0.87 & 0.82 & 0.59 & 0.65\\
%         Ours & \textbf{0.95} & \textbf{0.84} & \textbf{0.66} & \textbf{0.71}\\
%         \hline
%         \multicolumn{5}{|c|}{\bf $\text{DVN}_{\text{top}}$}\\
%         \hline
%         Random & 0.89 & 0.92 & 0.64 & 0.82\\ 
%         Coordinate & 0.89 & 0.93 & 0.74 & 0.78\\
%         Ours & \textbf{0.99} & \textbf{0.93} & \textbf{0.82} & \textbf{0.83}\\
%         \hline
%     \end{tabular}
% \label{tab:quantitative}
% \vspace{-3mm}
% \end{table}

\begin{table}
\centering
\caption{Quantitave comparison of our method with random and coordinate axes directions in terms of RCA and individual directions interpretability.}
\vspace{1mm}
    \begin{tabular}{ |c|c|c|c|c| }
        \hline
        Directions & MNIST & Anime & CelebA & BigGAN\\
        \hline
        \multicolumn{5}{|c|}{\bf Reconstructor classification accuracy}\\
        \hline
        Random & 0.46 & 0.85 & 0.6 & 0.76\\ 
        Coordinate & 0.48 & 0.89 & 0.82 & 0.66\\
        Ours & \textbf{0.88} & \textbf{0.99} & \textbf{0.9} & \textbf{0.85}\\
        \hline
        \multicolumn{5}{|c|}{\bf Individual interpretability (mean-opinion-score)}\\
        \hline
        Random & 0.06 & 0.00 & 0.29 & 0.19\\ 
        Coordinate & 0.00 & 0.00 & 0.22 & 0.66\\
        Ours & \textbf{0.47} & \textbf{0.26} & \textbf{0.30} & \textbf{0.69}\\
        \hline
    \end{tabular}
\label{tab:quantitative}
% \vspace{-3mm}
\end{table}

\subsection{Experiments on the ECSSD dataset}

We evaluate the described method on the ECSSD dataset \cite{ecssd}, which is a standard benchmark for weakly-supervised saliency detection. The dataset has separate train and test subsets, and we obtain the subset of classes $\mathcal{C}_\mathcal{D}$ from the train subset and evaluate on the test subset. %Here, we treat the out segmentation mask as the pixels with saliency 1 and out of mask as pixels with saliency 0.
For the segmentation model $U$, we take a simple U-net architecture \cite{unet}. We train $U$ on the pseudo-labeled dataset with Adam optimizer and the per-pixel cross-entropy loss with the temperature $10.0$. We perform 15000 steps with the initial rate of $0.005$ and decrease it by $0.2$ every 4000 steps and a batch size equal to 128. During inference, we rescale an input image to have a size 128 along its shorter side.

\begin{figure}
    \centering
    \includegraphics[width=0.49\textwidth]{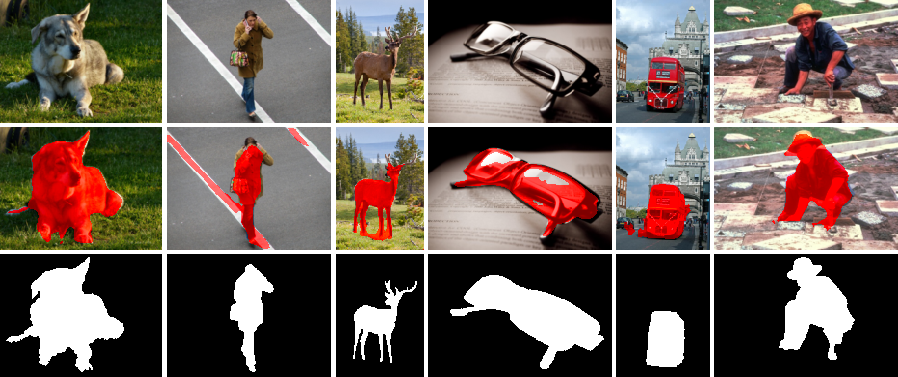}
    \vspace{-5mm}
    \caption{Results of saliency detection provided by our method. \textit{Line 1:} ECSSD images; \textit{Line 2:} predicted masks; \textit{Line 3:} groundtruth masks.}
    \vspace{-3mm}
    \label{fig:segmentation}
\end{figure}

We measure the model performance in terms of the mean average error (MAE), which is an established metric for weakly-supervised saliency detection. For an image $x$ and a groundtruth mask $m$, MAE is defined as:

\begin{equation}
\vspace{-3mm}
\text{MAE}(U(x), m) = \frac{1}{W \cdot H}\sum\limits_{i, j}|U(x)_{ij} - m_{ij})|
\end{equation}

where $H$ and $W$ are the image sizes. Our method based on BigGAN achieves $\text{MAE}$ equal to \textbf{$0.099$}, which is a competitive performance on ECSSD across the methods using the same amount of supervision \cite{wang2019salient} (i.e., image-level class labels from the ILSVRC dataset). \fig{segmentation} demonstrates several examples of saliency detection, provided by our method.

\section{Ablation}
\label{sect:ablation}

Here we present an ablation of the number of latent directions $K$ and the shift loss term. We ablate $K$ on MNIST and ILSVRC, see the results in \tab{mnist_K} and \tab{ilsvrc_K} in terms of individual interpretability (MOS) and RCA (see \sect{experiments} for metrics details). For each $K$ we also report the total number of interpretable directions according to the human evaluation. Notably, small values of $K$ are inferior since the classification task becomes easier and the model does not enforce directions to be ``disentangled''. On the other hand, higher $K$ does not harm interpretability but often results in duplicate directions.

\begin{table}[h]
    \centering
    \caption{Number of directions $K$ ablation for Spectral Norm GAN pretrained on MNIST dataset.}
    \begin{tabular}{|c||c|c|c|c|}
        \hline
         metrics & $K$ = 16 & 32 & 64 & 128 \\
         \hline
         MOS & 0.5 & 0.58 & 0.47 & 0.46 \\
         \hline
         MOS (absolute) & 8 & 19 & 30 & 59 \\
         \hline
         RCA & 0.98 & 0.95 & 0.88 & 0.79 \\
         \hline
    \end{tabular}
    \label{tab:mnist_K}
\end{table}

\begin{table}[h]
    \centering
    \caption{Number of directions $K$ ablation for BigGAN.}
    \begin{tabular}{|c||c|c|c|c|c|}
        \hline
         metrics & K = 15 & 30 & 60	& 90 & 120 \\
         \hline
         MOS & 0.3	& 0.3 & 0.38 & 0.75 & 0.69 \\
         \hline
         MOS (absolute) & 5 & 9 & 23 & 68 & 83 \\
         \hline
         RCA & 0.99 & 0.98 & 0.92 & 0.9 & 0.85 \\
         \hline
    \end{tabular}
    \label{tab:ilsvrc_K}
\end{table}

We also perform ablation of the shift loss term of the reconstructor $R$ by varying its multiplier. The ablation results for MNIST are presented in \tab{ablation_alpha}. Notably, the extreme values of the scaling factor $\lambda$ lead to quality degradation. In particular, $\lambda = 0$ leads to ``collapse'' directions, see \fig{collapse}. With high lambda the directions mostly become similar (e.g. all perform zoom).

\begin{table}[h]
    \centering
    \caption{Number of directions $K$ ablation for Spectral Norm GAN pretrained on MNIST dataset.}
    \begin{tabular}{|c||c|c|c|c|c|}
        \hline
         metrics & $\lambda$ = 0 & 0.125 & 0.25 & 0.5 & 2\\
         \hline
         MOS & 0.27 & 0.35 & 0.47 & 0.42 & 0.25 \\
         \hline
         RCA & 0.88 & 0.90 & 0.88 & 0.87 & 0.75 \\
         \hline
    \end{tabular}
    \label{tab:ablation_alpha}
\end{table}

\section{Conclusion}
\label{sect:conclusion}

In this paper, we have addressed the discovery of interpretable directions in the GAN latent space, which is an important step to an understanding of generative models required for researchers and practitioners. Unlike existing techniques, we have proposed a completely unsupervised method, which can be universally applied to any pretrained generator. On several standard datasets, our method reveals interpretable directions that have never been observed before or require expensive supervision to be identified. Finally, we have shown that one of the revealed directions can be used to generate high-quality synthetic data for the challenging problem of weakly supervised saliency detection. We expect that other interpretable directions can also be used to improve the performance of machine learning in existing computer vision tasks.
\bibliographystyle{icml2020}
\bibliography{main}

\onecolumn

\section{Supplementary material}

\subsection{Orthogonal parametrisation}
\label{sect:ortho_sup}
In this section we present the details on the orthonormal matrix parametrisation we use.  Let us define as $\mathrm{Skew}_d$ the set of all real $d \times d$ skew-symmetric matrices $\{S\ |\ S^T = -S\}$. We also write $\mathrm{Mat}_{d \times d}$ for the set of all real $d \times d$ matrices. Clearly, $\mathrm{Skew}_d$ can be associated with a $\frac{d \cdot (d - 1)}{2}$-dimensional space as it is uniquely defined by the elements under the diagonal. One can consider the exponential map $\mathrm{exp}: \mathrm{Mat}_{d \times d} \to \mathrm{Mat}_{d \times d}$ that can be defined explicitly as $\mathrm{exp}(A) = \sum\limits_{n = 0}^{\infty} \frac{A^n}{n!}$. It is known that this sum always converges and defines a bijective smooth map from $\mathrm{Skew}_d$ to the space of all orthogonal matrices with a positive determinant $SO(d)$. Once we are looking for non-oriented interpretable directions, without loss of generality we may assume that the desired latent basis has positive orientaion. Otherwise we may flip the first direction. Following these considerations, we may use $\mathbb{R}^\frac{d \cdot (d - 1)}{2}$ as the latent directions parametrisation once we set them to be orthonormal. See e.g. \citet{fulton2013representation} for further details.

\subsection{Direction Variation Naturalness (DVN)}
\label{sect:dvn_sup}

\begin{figure}[h!]
    \centering
    \includegraphics[width=0.85\textwidth]{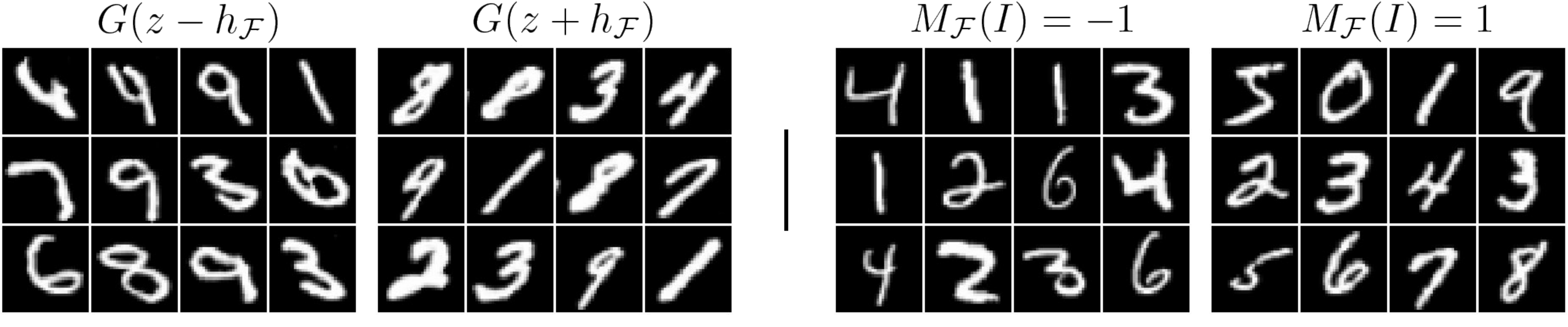}
    \caption{\textit{Left}: generator samples from $\mathcal{D}_\mathcal{F}$ grouped by latent shift direction. \textit{Right}: split of the real MNIST images according to a model, trained on the generated samples split. They form the pseudo-labeled dataset $\mathcal{D}^{\mathrm{real}}_\mathcal{F}$}.
    \label{fig:mnist_pseudolabels}
\end{figure}

\begin{figure}
    \centering
    \includegraphics[width=0.9\textwidth]{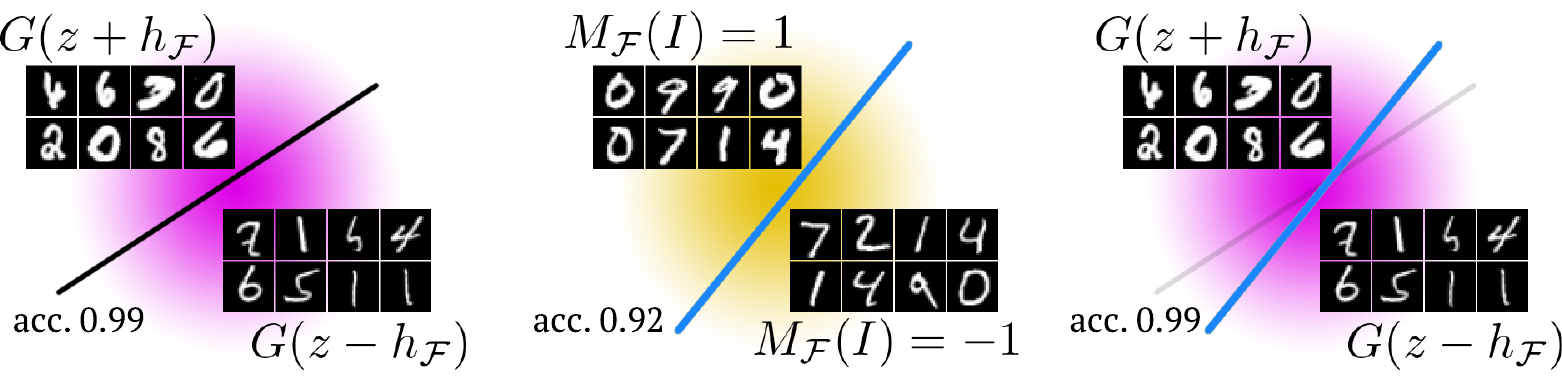}
    \vspace{2mm}

    \rule{4cm}{0.5pt}
    \vspace{2mm}

    \includegraphics[width=0.8\textwidth]{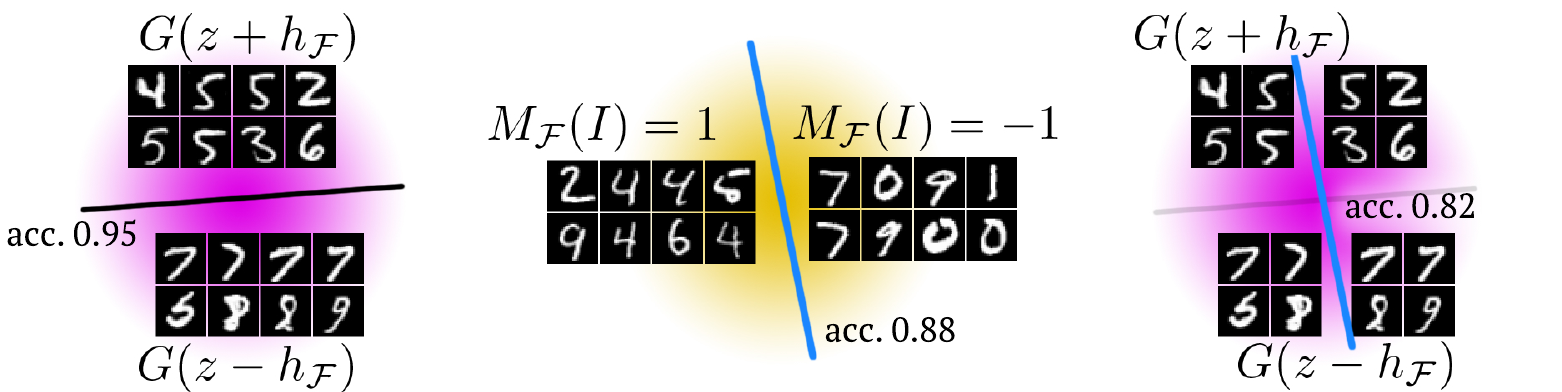}
    \caption{DVN computation process. \textit{Purple}: generated images domain, \textit{yellow}: real images domain. \textit{Top row}: the split of generated images naturally transfers to the real images domain and finally induces almost the same split of the generated images. \textit{Bottom row}: the split of the generated images is difficult to interpret and does not correspond to any natural factor of variation. Therefore, it is hard for a simple classification model to ``generalize'' to real data, which results in lower DVN values.}
    \label{fig:dvn_splits}
\end{figure}

We also experimented with an alternative measure for individual interpretability that does not require human supervision. We refer to this measure as Direction Variation Naturalness (DVN). DVN measures how ``natural'' is a variation of images obtained by moving in a particular direction in the latent space. Intuitively, a natural factor of variation should appear in both real and generated images. Furthermore, if one splits the images based on the large/small values of this factor, the splitting should operate similarly for real and generated data.
We formalize this intuition as follows. Let us have a direction $h_\mathcal{F}$ in the latent space that corresponds to a semantic factor $\mathcal{F}$. We always scale the shift $h_\mathcal{F}$ length to be equal to 6 which is the maximal amplitude while training. Then we can construct a pseudo-labeled dataset for binary classification $\mathcal{D}_\mathcal{F} = \{(G(z \pm h_\mathcal{F}), \pm 1)\}$ with $z \sim \mathcal{N}\left(0,I\right)$. Given this dataset, we train a binary classification model $M_\mathcal{F}: G(z) \longrightarrow \{-1, 1\}$ to fit $\mathcal{D}_\mathcal{F}$. After that, $M_\mathcal{F}$ can induce pseudolabels for the dataset of real images $\mathcal{D}$, which results in pseudo-labeled dataset 
$\mathcal{D}_\mathcal{F}^{real}{=}\{(I, M_\mathcal{F}(I)), I \in \mathcal{D}\}$ (see \fig{mnist_pseudolabels}).
We expect that if the factor of variation $\mathcal{F}$ is responsible for a single and easy-to-interpret attribute, the split of real images $\mathcal{D}_\mathcal{F}^{real}$ will be consistent with $\mathcal{D}_\mathcal{F}$. Therefore, the pseudo-labels-pretrained classifier is expected to demonstrate high performance on $\mathcal{D}_\mathcal{F}$. On the contrary, if the factor of variation is mixed and uninterpretable, we expect that the classifier, trained on $\mathcal{D}^{\mathrm{real}}_\mathcal{F}$, will perform poorly on $\mathcal{D}_\mathcal{F}$ (see \fig{dvn_splits}).
Thus, we re-train the model $M_\mathcal{F}$ from scratch on $\mathcal{D}_\mathcal{F}^{real}$ and compute its accuracy on $\mathcal{D}_\mathcal{F}$. The obtained accuracy value is referred to as DVN.

In the experiments below, we report the DVN averaged over all directions. Since the directions with higher DVN values are typically more interpretable, we additionally report the average DVN over the top $50$ directions ($\text{DVN}_{\text{top}}$). Following the experiment in \sect{experiments}, in \tab{quantitative_dvn} the directions discovered by our method are compared in terms of DVN with random and coordinate directions.

In all the experiments, we use a LeNet-like classification model (see \tab{dvn_model}) with the cross-entropy objective. We train it for both $\mathcal{D}_\mathcal{F}$ and $\mathcal{D}^{\mathrm{real}}_\mathcal{F}$ for $100$ steps of Adam optimizer, as in all our experiments it converges rapidly. We use batch $32$ and learning rate $0.001$. The sizes of $\mathcal{D}_\mathcal{F}$ and $\mathcal{D}^{\mathrm{real}}_\mathcal{F}$ always equal $3200$ and we did not observe any difference from the usage of more samples.% Following \cite{beta-VAE} we also experimented with a simple linear classifier yet it commonly demonstrated accuracy close to $0.5$ on complicated domains. 

\begin{figure}[h!]
    \centering
    \includegraphics[width=0.85\textwidth]{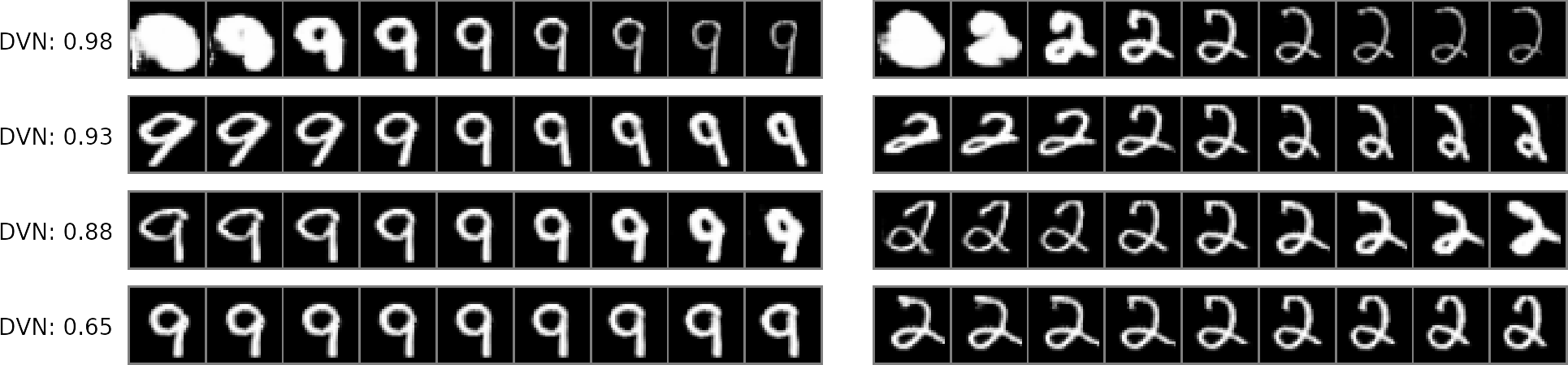}
    \caption{Image variations obtained by moving latent codes along four directions and the corresponding DVN values. The directions with high DVN are easier to interpret.}
    \label{fig:dvn_samples}
\end{figure}

\begin{table}
\centering
\caption{Quantitave comparison of our method with random and coordinate axes directions in terms of DVN.}
\vspace{1mm}
    \begin{tabular}{ |c|c|c|c|c| }
        \hline
        Directions & MNIST & Anime & CelebA & ILSVRC\\
        \hline
        \multicolumn{5}{|c|}{\bf DVN}\\
        \hline
        Random & 0.87 & 0.81 & 0.56 & 0.71\\ 
        Coordinate & 0.87 & 0.82 & 0.59 & 0.65\\
        Ours & \textbf{0.95} & \textbf{0.84} & \textbf{0.66} & \textbf{0.71}\\
        \hline
        \multicolumn{5}{|c|}{\bf $\text{DVN}_{\text{top}}$}\\
        \hline
        Random & 0.89 & 0.92 & 0.64 & 0.82\\ 
        Coordinate & 0.89 & 0.93 & 0.74 & 0.78\\
        Ours & \textbf{0.99} & \textbf{0.93} & \textbf{0.82} & \textbf{0.83}\\
        \hline
    \end{tabular}
\label{tab:quantitative_dvn}
\end{table}

\begin{table}[!h]
\centering
\renewcommand{\arraystretch}{1.2}
\begin{tabular}{c}
\hline \hline
\textbf{LeNet-based binary classifier} \\
\hline
input: $C \times h \times w$\\\hline
\textsc{Conv}, kernel: $5\times 5$, channels: $6$\\\hline
\textsc{BN, ReLU}\\\hline
\textsc{MaxPool}, kernel: $2 \times 2$, stride: $2$\\\hline
\textsc{Conv}, kernel: $5\times 5$, channels: $16$\\\hline
\textsc{BN, ReLU}\\\hline
\textsc{MaxPool}, kernel: $2 \times 2$, stride: $2$\\\hline
\textsc{Conv}, kernel: $5\times 5$, channels: $120$\\\hline
\textsc{BN, ReLU, AvgPool}\\\hline
\textsc{FC}, channels: $84$\\
\textsc{BN, ReLU}\\
\textsc{FC}, channels: $2$\\
\hline \hline
\end{tabular}
\renewcommand{\arraystretch}{1}
\caption{The binary classification model used for DVN computation.}
\label{tab:dvn_model}
\end{table}

\subsection{Discovered directions uniformity.}
As for further analysis, we report the global affect of a latent shift for some of the discovered directions. We calculate the Fr\'echet Inception Distance \cite{fulton2013representation} between the real images and the shifted distribution $\{G(z + h), z \sim \mathcal{N}(0, I)\}$ for different shift magnitudes. We compute FID with ILSVRC test data and 50.000 randomly sampled latent $z$. \tab{fid} presents the affect of different shift scales for some of directions from BigGAN latent space. The base model performs with FID = $10.2$. Notably, the generated data transformation appears to be more homogeneous for different directions compared to i.e. \cite{jahanian2019steerability}. Apparently, this happens because we learn these directions simultaneously instead of independently. 

\begin{table}[h!]
    \centering
    \begin{tabular}{|c||c|c|c|c|c|c|c|}
        \hline
         shift scale & -12 & -6 & -3 & 0 & 3 & 6 & 12 \\
        \hline
         Backgndound removal & 16.5  & 11.8 & 10.4 & 10.2 & 11.3 & 14.9 & 27.3 \\
         Lighting  & 13.0 & 11.0 & 10.3 & 10.2 & 10.6 & 11.4 & 15.3 \\
         Vertical shift & 40.3 & 15.1 & 12.7 & 10.2 & 11.8 & 14.3 & 44.2 \\
         Zooming & 21.6 & 12.6 & 10.9 & 10.2 & 10.4 & 11.8 & 17.9 \\
         \hline
    \end{tabular}
    \caption{FID for different shift magnitudes for some of explored latent directions.}
    \label{tab:fid}
\end{table}

\subsection{Alternative disentanglement metrics.}

In our work, we have introduced three quantitative measures to compare different sets of directions from the same latent space. We do not use the common disentanglement metric from the $\beta$-VAE paper \cite{beta-VAE}, since it heavily relies on an additional encoder that can be difficult to obtain for existing GAN models. Moreover, metric from \cite{beta-VAE} is typically applied for a relatively small number of factors $K$ (up to five) and it is unclear if it is reliable for large $K$.

\subsection{Other details}
\begin{figure}[h]
    \centering
    \includegraphics[width=0.8\textwidth]{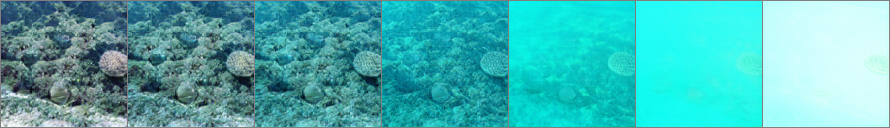}
    \caption{Variation along the background removal direction for the BigGAN generator with class ``coral reef''. As there seems to be no foreground, the model blanks the whole image}
    \label{fig:no_bg}
\end{figure}

Interestingly, the discovered directions sometimes behave in an unexpected manner. For instance, we have observed that the BigGAN direction responsible for the background removal simply blanks the images that do not have explicit foreground objects (see \fig{no_bg}).

As a final comment, we describe the exact procedure we have followed to find the desired interpretable directions. Once the method proposes $K$ directions, we sort them with respect to the individual DVN values. Then for each direction $h_k$, we draw the images $G(z + \varepsilon A(h_k))$ varying $\varepsilon$ from $-9$ to $9$. We review them manually and highlight the most interesting directions. For instance,  for $K{=}128$ this procedure takes about ten minutes for a single person.

% \bibliographystyleSupmat{plain}
% \bibliographySupmat{main}

\end{document}